\documentclass[
  journal=medium,
  manuscript=article-type,
  year=2020,
  volume=37,
]{cup-journal}

\usepackage{xpatch}

\makeatletter

\xpatchcmd{\@maketitle}
  {(\cup@year), {\volumefont\cup@vol}, \thepage--\pageref{LastPage}}
  {}
  {}{}

\xpatchcmd{\@maketitle}
  {{\sffamily\bfseries\color{structure@color}\MakeUppercase{\convertchar{\cup@manuscript}{-}{ }}\par}}
  {}
  {}{}

\xpatchcmd{\@maketitle}
  {\hfill\includegraphics[width=26mm]{}\par}
  {\par}
  {}{}

\makeatother

\usepackage{amsmath}
\usepackage{amssymb}
\usepackage[nopatch]{microtype}
\usepackage{booktabs}
\newcommand{\institution}[1]{#1}
\usepackage{booktabs}      
\usepackage{tabularx}      
\usepackage{array}         
\usepackage{multirow}      
\usepackage{graphicx}      
\newcolumntype{N}{>{\centering\arraybackslash}p{0.75cm}} 


\makeatletter
\newenvironment{Example}{
  \begin{mdframed}[
    innerlinewidth=0.5pt,
    innerleftmargin=9pt,
    innerrightmargin=9pt,
    innertopmargin = 10pt,
    innerbottommargin=10pt,
    skipabove=\dimexpr\topsep+\ht\strutbox\relax,
    roundcorner=5pt,
    frametitle={},          
    frametitlerule=false    
  ]
}{
  \end{mdframed}
}
\makeatother

\usepackage{microtype}
\usepackage{hyperref}
\usepackage{url}
\usepackage{booktabs}

\usepackage{xspace}
\usepackage{multirow}   
\usepackage{graphicx} 

\usepackage{tikz}
\usetikzlibrary{arrows.meta, positioning, shapes.geometric}

\newcommand{\gemma}{\textsc{Gemma-3-4B}\xspace}
\newcommand{\qwen}{\textsc{Qwen3-4B}\xspace}
\newcommand{\llama}{\textsc{Llama-3.2-3B}\xspace}
\newcommand{\cometkiwi}{\textsc{CometKiwi}\xspace}
\newcommand{\deepseek}{\textsc{DeepSeek-V3}\xspace}
\newcommand{\xcomet}{\textsc{xCOMET}\xspace}
\newcommand{\gemini}{\textsc{Gemini}\xspace}

\newcommand{\aloperl}{\texttt{ALOPE-RL}\xspace}
\newcommand{\tqr}{\textsc{TQR}\xspace}
\newcommand{\wtag}{\textsc{WTag}\xspace}
\newcommand{\allrewards}{\textit{All rewards}\xspace}
\newcommand{\corerewards}{\textit{Core rewards}\xspace}
\newcommand{\daonly}{\textit{DA-only rewards}\xspace}
\newcommand{\equalweightedmean}{\textit{Equal-weighting}\xspace}
\newcommand{\weightedsum}{\textit{Weighted sum}\xspace}

\usepackage{makecell}      
\usepackage{adjustbox}
\newcommand{\country}[1]{#1}

\title{Beyond Scalar Scores: Reinforcement Learning for Error-Aware Quality Estimation of Machine Translation} 

\author{Archchana Sindhujan}
\affiliation{%
  \institution{Institute for People-Centred AI},
  \institution{University of Surrey},
  \country{UK}
}
\email{a.sindhujan@surrey.ac.uk}

\author{Girish A. Koushik}
\affiliation{%
  \institution{NICE Research Group},
  \institution{University of Surrey},
  \country{UK}
}

\author{Shenbin Qian}
\affiliation{%
  \institution{Department of Informatics},
  \institution{University of Oslo},
  \country{Norway}
}

\author{Diptesh Kanojia}
\affiliation{%
  \institution{Institute for People-Centred AI},
  \institution{University of Surrey},
  \country{UK}
}

\author{Constantin Or\u{a}san}
\affiliation{%
  \institution{Centre for Translation Studies},
  \institution{University of Surrey},
  \country{UK}
}


\begin{document}
\begin{abstract}
Quality Estimation (QE) aims to assess the quality of machine translation (MT) outputs without relying on reference translations, making it essential for real-world, large-scale MT evaluation. Large Language Models (LLMs) have shown significant promise in advancing the field of quality estimation of machine translation. However, most of the QE approaches solely rely on scalar quality scores, offering no explicit information about the translation errors that \textit{should} drive these judgments. Moreover, for low-resource languages where annotated QE data is limited, existing approaches struggle to achieve reliable performance. To address these challenges, we introduce the first segment-level QE dataset for English$\rightarrow$Malayalam, a severely resource-scarce language pair in the QE domain, comprising human-annotated Direct Assessment (DA) scores and Translation Quality Remarks (TQR), which are short, contextual, free-form annotator comments that describe translation errors. We further introduce \aloperl, a policy-based reinforcement learning framework that trains efficient adapters based on policy rewards derived from DA score and \tqr.  Integrating error-aware rewards with \aloperl, enables LLMs to reason about translation quality beyond numeric scores. Despite being trained on a small-scale QE dataset, \aloperl achieves state-of-the-art performance on English$\rightarrow$Malayalam QE using compact LLMs ($\leq$4B parameters) fine-tuned with LoRA and 4-bit quantization, outperforming both larger LLM-based baselines and leading encoder-based QE models. Our results demonstrate that error-aware, policy-based learning can deliver strong QE performance under limited data and compute budgets. We release our dataset, code, and trained models to support future research\footnote{\href{xxxxx}{Link to this repository will be published upon acceptance}}.
\end{abstract}

\section{Introduction}
In recent years, large language models (LLMs) have significantly advanced machine translation (MT) evaluation methods~\autocite{gain2025bridginglinguisticdividesurvey,kocmi2025preliminaryrankingwmt25general}, including the field of Quality Estimation (QE), which assesses MT quality without reference translations~\autocite{info16100916}. Recent findings~\autocite{lavie-etal-2025-findings} reveal that although LLM-based evaluators perform well for high-resource languages~\autocite{junczys-dowmunt-2025-gemba,maheswaran-etal-2025-taser}, they exhibit substantially lower correlation with human judgments for low-resource languages~\autocite{sindhujan-etal-2025-alope}. These findings highlight a consistent challenge for language models, which struggle to perform reliable QE without contextual supervision in low-resource settings.

Conventional Direct Assessment (DA) score ~\autocite{graham-etal-2013-continuous} based QE systems take only the source segment and machine-translated hypothesis as input and learn to predict a single scalar score which indicates the quality of the MT system ($0$–$100$). However, because DA offers no contextual insight into the underlying translation errors that shaped the judgment, such systems learn quality as an undifferentiated signal, often \textit{lacking the necessary interpretation for error reasoning}. Fine-grained annotation metrics such as \href{https://themqm.org/error-types-2/typology/}{Multi-Dimensional Quality Metric} (MQM;~\citealt{lommel-etal-2014-using}) or Error Span Annotation (ESA;~\citealt{kocmi-etal-2024-error}) offer additional error context and supervision through span-level error labels and severity judgments. Although ESA is lighter than MQM, both frameworks still require substantial annotator expertise, span selection, and challenging annotation, making them costly and difficult to scale. This reinforces the need for simpler yet context-bearing annotation signals that capture translation errors without imposing high cognitive load or time demands on annotators.

\begin{figure*}[t]
 \centering
 \includegraphics[width=0.99\textwidth, keepaspectratio]{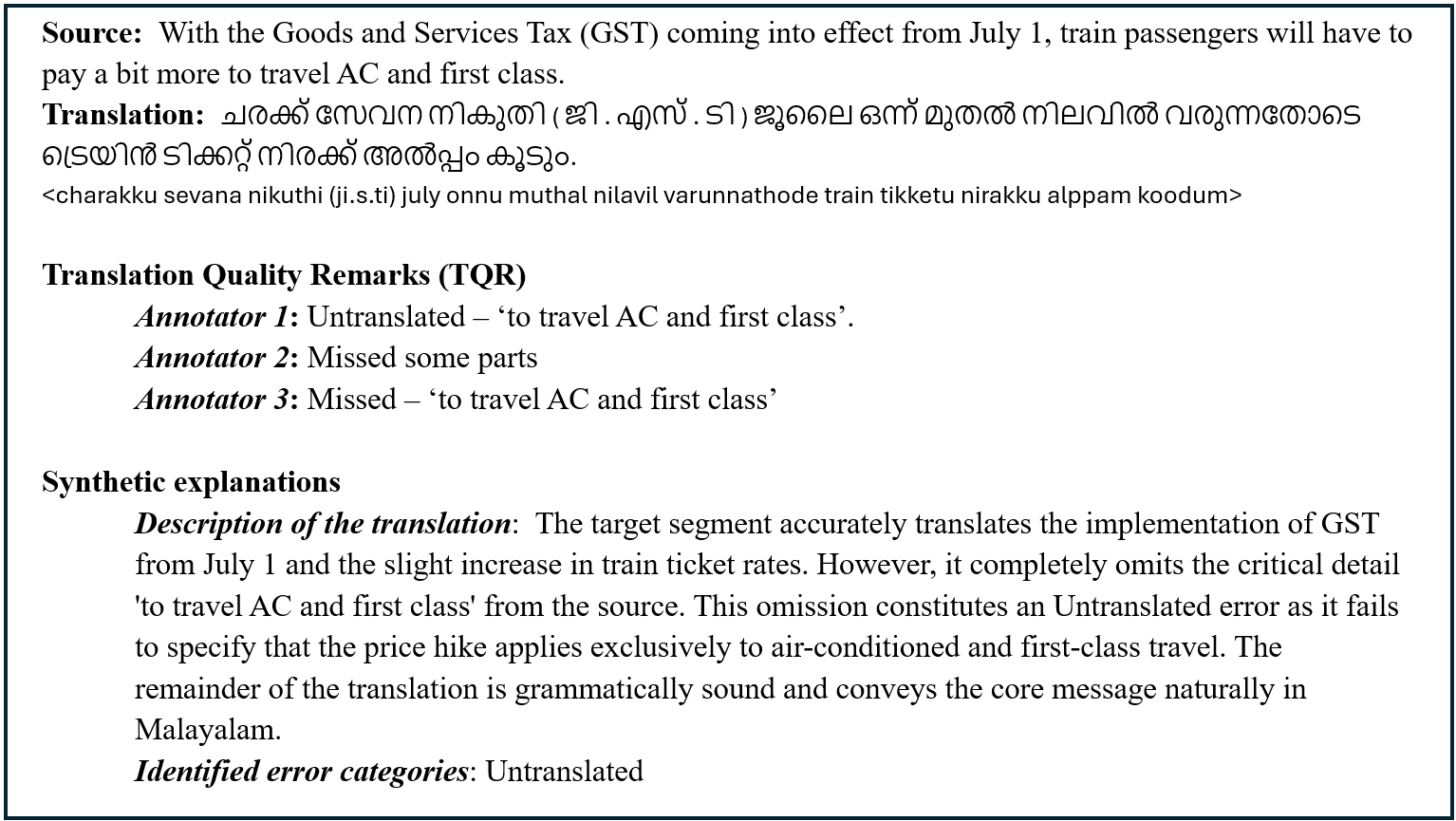}
\caption{ Example of the human annotated \tqr and the generated synthetic explanations}
\label{fig:example} 
\end{figure*}

Considering these limitations, we introduce the first segment-level quality estimation dataset for English$\rightarrow$Malayalam (En$\rightarrow$Ml), an extremely low-resource language pair for which no prior QE datasets exist. Our dataset includes DA scores along with Translation Quality Remarks (\tqr) (see Figure~\ref{fig:example}). We propose TQR as a short-form comment in natural language, indicating the error type and/or erroneous spans from either the source or MT hypothesis. We posit that this free-form short comment provides sufficient error context to help QE systems shape DA judgement. Annotators were provided with additional guidelines on error types but were allowed to describe errors freely, without relying on fixed taxonomies or detailed span annotations. In contrast to MQM’s extensive span marking and complex taxonomies, and ESA’s continued reliance on span identification and structured labels, \tqr offers a simple and more flexible annotation signal. This design provides rich contextual information while keeping annotation time and cognitive load manageable ($\S$~\ref{subsec: dataset}). 

We treat \tqr as a weak supervision signal that enables contextual reasoning about translation errors beyond what can be captured through scalar DA scores alone. Prior work on Reinforcement-Learning (RL) shows that even short, free-form natural-language signals can serve as effective information for LLMs ~\autocite{zhang2025critiquegrpoadvancingllmreasoning}, indicating the possibility of \tqr to enhance QE performance under RL.

RL approaches have recently been explored for MT~\autocite{he-etal-2024-improving,uhlig-etal-2025-cross,ramos2025finegrainedrewardoptimizationmachine,feng2025mtr1zeroadvancingllmbasedmachine}, yet their application to MT evaluation remains comparatively limited. \citet{tan-monz-2025-remedy} propose an approach for MT evaluation with reward-modeling from learned pairwise human preferences, demonstrating the potential of preference-based optimization for metric learning. However, this method has primarily been evaluated on high-resource language pairs and depends on explicit preference annotations, revealing a gap in RL-based MT evaluation approaches that can operate effectively with natural-language signals.

Motivated by this gap, we propose \aloperl\footnote{ALOPE-RL - \textbf{A}daptive \textbf{L}earning and \textbf{O}ptimization for \textbf{P}olicy-based quality \textbf{E}stimation with \textbf{R}einforcement \textbf{L}earning.}, a policy-driven reinforcement-learning framework that adapts LLMs for QE via multi-reward optimization. We train the models with Group Relative Policy Optimization (GRPO) ~\autocite{shao2024deepseekmathpushinglimitsmathematical} and utilize \tqr as a supervision signal, allowing the model to reason about the underlying causes of translation errors instead of relying solely on scalar DA scores. For policy-based learning, we also leverage \tqr to generate synthetic explanations (Figure~\ref{fig:example}) comprising `Identified error categories' and `Description of the translation' which act as reference signals for reward components, helping the model score accurately with interpretable error attribution. To ensure efficiency in low-resource settings, we train on small annotated datasets ($\leq$5K) and compact LLMs ($\leq$4B parameters) fine-tuned with LoRA~\autocite{hu2021lora} and 4-bit quantization~\autocite{dettmers2023qloraefficientfinetuningquantized}, yielding a lightweight yet robust QE system. 
\\
Our main contributions are as follows:
\begin{itemize}
    \item We introduce the first segment-level QE dataset for the English$\rightarrow$Malayalam language pair, containing human-annotated Direct Assessment scores and \textit{Translation Quality Remarks (\tqr)} for error-aware QE. 
    
    \item We demonstrate that incorporating \tqr as contextual input substantially improves LLM-based QE performance, demonstrating the value of weak human annotations for enhancing translation-error reasoning.

    \item We introduce \aloperl, a policy-based reinforcement-learning framework that leverages \tqr-driven weak supervision for QE and achieves state-of-the-art performance under low-resource conditions with smaller LLMs, outperforming both leading encoder-based approaches and strong LLM baselines.

\end{itemize}

\section{Background}
\textbf{Encoder-based QE.} Transformer-based multilingual encoders leverage pretrained cross-lingual representations for regression-based QE~\autocite{blain-etal-2023-findings,rei-etal-2022-cometkiwi,ranasinghe-etal-2020-transquest,perrella2022matese}. Models such as \cometkiwi~\autocite{rei-etal-2023-scaling} and \xcomet~\autocite{guerreiro-etal-2024-xcomet} are the current state-of-the-art approaches. \cometkiwi integrates multi-task learning with both regression and classification losses to enhance interpretability and accuracy, while \xcomet further incorporates MQM ~\autocite{burchardt-2013-multidimensional}. These models have been trained with an extensive amount of QE data.

\textbf{LLM-based QE.} Large language models have recently been investigated as metrics for translation quality estimation, with several studies demonstrating strong results on high-resource language pairs. GEMBA~\autocite{kocmi-federmann-2023-large} shows that GPT-based evaluators can achieve competitive correlations with human judgments, compared to reference-based MT evaluation. Similar findings are reported with TASER~\autocite{maheswaran-etal-2025-taser} and GEMBA-V2~\autocite{junczys-dowmunt-2025-gemba}. However, these promising results are largely confined to high-resource settings. Evidence from prior work on low-resource QE~\autocite{vandan-etal-2023-towards,sindhujan-etal-2025-llms} indicates that LLMs, whether prompted or fine-tuned, struggle to obtain strong correlation with human judgment for QE, and often underperform compared to encoder-based metrics such as COMETKiwi~\autocite{rei-etal-2023-scaling}. Recent results from the WMT25 Shared Task on Automated Translation Evaluation~\autocite{lavie-etal-2025-findings} further reinforce this disparity, highlighting the limitations of current LLMs in achieving strong QE performance under resource-scarce conditions.

\section{Methodology}
\subsection{Dataset} \label{subsec: dataset}
Malayalam remains an under-resourced language, with limited digitisation and parallel corpora for machine translation, with no publicly available data for reference-less MT evaluation of English to Malayalam. In this work, we present the first Direct Assessment score–based QE dataset for the En$\rightarrow$Ml language pair, comprising $5$K instances in total drawn from the Finance, News, and Legal domains extracted from the Anuvaad corpus \footnote{\url{https://github.com/project-anuvaad/anuvaad-parallel-corpus}} and translated with the IndiTrans2 MT model ~\autocite{gala2023indictrans2highqualityaccessiblemachine}. Each source-translation pair contains DA score judgments from three human annotators. For DA annotation, we adhered to the FLORES guidelines~\autocite{guzman2019flores}. Unlike previous DA-based QE datasets, this resource incorporates Translation Quality Remarks (\tqr) from human annotators for the source–translation pair (~\ref{app:DA_guideline}).  

\subsubsection{Translation Quality Remarks (\tqr)}
Alongside assigning DA scores, annotators were given guidelines (~\ref{app:tqr-guideline}) outlining the types of errors Untranslated, Addition, Mistranslation, Fluency Error, Other) observed in the translations to provide brief remarks. These \tqr (Figure~\ref{fig:example}) serve as \textit{weak human annotations}, enriching DA supervision with lightweight, interpretable signals about errors in the translation. \tqr provides practical alternative to MQM-style annotations, which require error span marking, error category selection, and severity labelling, making them time-consuming, costly, and challenging to scale. 

\begin{figure}[t]
\centering
\scriptsize
\begin{Example}{} 

You are an expert in identifying the errors in the translations of English to <TARGET LANGUAGE>. The target language is morphologically complex, and in most cases the errors in the machine-translated target segment can be categorized into the following categories:\\

    \textit{Untranslated}: A word or phrase in the source is omitted in the translation.\\
    \textit{Addition}: The translation includes a word or phrase not present in the source.\\
    \textit{Mistranslation}: A word or phrase in the translation does not accurately represent the source meaning.\\
    \textit{Fluency Error}: The translation sounds unnatural due to grammar, spelling, punctuation, or inconsistency.\\
    \textit{Other}: Any other error not covered by the above categories.\\
If there are no errors in the translation, assign the error category as ‘\textit{No Errors}’.\\
    
Your task is to provide a detailed natural language description about the translation discussing any errors within the target segment in 100 words. Use the error categories above and indicate all errors in the translation. \\

The English source segment and the Malayalam target segment are provided, along with translation quality remarks which comment on the translation and any identified errors (if present). \\

Following are some examples of detailed natural language description of the translation.\\
\textbf{Example 1: }\\
Source segment: <Source sentence>\\
Target segment: <Translated sentence>\\
Translation Quality Remark : (<Remark 1>, <Remark 2>, <Remark 3>)\\
Detailed description of the translation: <Example description from human annotator>\\

\textbf{Example 2: }\\
Source segment: <Source sentence>\\
Target segment: <Translated sentence>\\
Translation Quality Remark : (<Remark 1>, <Remark 2>, <Remark 3>)\\
Detailed description of the translation: <Example description from human annotator>\\

\textbf{Example N: }\\
...\\
\\
The English source segment and the Malayalam target segment need to be analysed are provided below along with the remarks.\\

\textbf{Source segment}: <Source sentence> \\
\textbf{Target segment}: <Translated sentence>\\
\textbf{Translation Quality Remark} : (<Remark 1>, <Remark 2>, <Remark 3>) \\

Analyse the translation according to the given instructions and return the description of the translation in the following JSON format only:\\
\{\{\\
  "\textbf{Description of the translation}": "",\\
  "\textbf{Identified error categories}": "",\\
\}\}

\end{Example}
\caption{Prompt template for synthetic data generation with Translation Quality Remarks.}\label{fig:syn_TQR}

\end{figure}

\textbf{Synthetic Explanations \textit{via} TQR.}  We leveraged the \tqr in the En$\rightarrow$Ml dataset, together with the corresponding source and translated segments, to generate synthetic explanations containing a detailed description of the translation and error categories (Figure~\ref{fig:example}). These synthetic outputs serve as \textit{reference signals} for several reward components in our experiments. To construct the synthetic explanations, we experimented with multiple prompting strategies and generated outputs using two strong reasoning models, \gemini-2.5-pro and \deepseek. An \textit{evaluation on $20$ instances, reviewed independently by two native Malayalam speakers}, was conducted to assess the clarity and correctness of the generated error descriptions. Based on their feedback, we selected the prompt presented in Figure~\ref{fig:syn_TQR} as it yielded the most accurate and interpretable explanations. Given that Malayalam evaluators rated the synthetic explanations from both models as comparable, with a marginal preference for \deepseek, we retain outputs from both models for use in the initial En$\rightarrow$Ml experiments.

\begin{table}[t]
\centering
\small
\setlength{\tabcolsep}{4pt}
\renewcommand{\arraystretch}{1.2}
\begin{tabular}{lcccc}
\toprule
\textbf{Data Split} & \textbf{En$\rightarrow$Ml\textsuperscript{\dag}} & \textbf{En$\rightarrow$Ta} & \textbf{En$\rightarrow$Mr} & \textbf{En$\rightarrow$Hi} \\
\midrule
Train & 4000 & 2145 & 4000 & 1499 \\
Test  & 1000 & 500  & 1000 & 500 \\
\bottomrule
\end{tabular}
\caption{Dataset size for each language pair. 
\textsuperscript{\dag}\,En$\rightarrow$Ml contains both \textbf{\tqr} and \textbf{Word-level QE Tags (\wtag)}; other language pairs include only \textbf{\wtag}.}
\label{tab:data_split}
\end{table}

\subsection{Other Language Pairs}\label{other_lang_synthetic_data}
In addition to the En$\rightarrow$Ml dataset, we extend our study to English$\rightarrow$Tamil (En$\rightarrow$Ta), English$\rightarrow$Marathi (En$\rightarrow$Mr), and English$\rightarrow$Hindi (En$\rightarrow$Hi), where only DA scores are available, and no human-annotated \tqr exist. Therefore, we derive weak supervision from tokenized word-level QE tags (\wtag)~\autocite{zerva-etal-2022-findings}, which indicate erroneous tokens via binary OK/BAD labels obtained through post-edit alignments. Both signals aim to capture translation errors, while \wtag identifies the erroneous words via token-level judgments, \tqr indicates erroneous words with brief natural language descriptions. This extension allows us to evaluate the generalizability of our framework across language pairs and compare the impact of \tqr \textit{vs.} \wtag as weak supervision signals. These language pairs are considered low-resourced for QE due to the limited size of labelled datasets, between 1.5K and 4K training instances per pair (Table~\ref{tab:data_split}), which are well below the scale needed for training strong QE models.

To obtain \wtag, we first identify DA-annotated instances for each language pair that overlap with the corresponding Post-Editing (PE) datasets from the WMT23 QE shared task, where professional translators revised MT outputs to produce post-edited reference translations~\autocite{blain-etal-2023-findings}. We derive \wtag for each overlapping instance following the annotation conventions of \citet{zerva-etal-2022-findings} and using Translation Edit Rate (TER)~\autocite{snover-etal-2006-study}. We align MT hypotheses with their post-edited counterparts and label each translated token as ‘OK’ (correct) or ‘BAD’ (incorrect). These token-level tags serve as weak supervision signals (instead of \tqr) and are incorporated into the prompt (~\ref{app:synthetic_prompts}-Figure~\ref{fig:syn_wor-level}) to generate synthetic explanations. For comparison, we also generate \wtag for the En$\rightarrow$Ml dataset~\footnote{We release the human post-edited En$\rightarrow$Ml dataset along with the QE data} following the same procedure. The details of the final data splits for all language pairs are provided in Table~\ref{tab:data_split}.

\begin{figure}[t]
\centering
\scriptsize
\begin{Example}{} 
\textbf{System:} \\
You are an expert in evaluating English to <TARGET LANGUAGE> machine translations. Your task is to provide a comprehensive evaluation, including a quality score, error categorization, and a detailed analysis.\\
    \textbf{\textit{Scoring Guidelines (0-100):}}\\
    1.  0-30: Mostly unintelligible - completely inaccurate or containing only some keywords.\\
    2.  31-50: Partial intelligibility - some keywords present but numerous grammatical errors. \\
    3. 51-70: Generally clear - most keywords included with only minor grammatical errors. \\
    4. 71-90: Clear and intelligible - all keywords present with only minor non-grammatical issues. \\
    5. 91-100: Perfect or near-perfect - accurately conveys source meaning without errors. \\
    \\
    \textbf{\textit{Error Categorization Guidelines:}}\\
    1.  Untranslated: A word or phrase in the source is omitted in the translation.\\
    2.  Addition: The translation includes a word or phrase not present in the source.\\
    3.  Mistranslation: A word or phrase in the translation does not accurately represent the source meaning.\\
    4.  Fluency Error: The translation sounds unnatural due to grammar, spelling, punctuation, or inconsistency.\\
    5.  Other: Any other error not covered by the above categories.\\
    6.  No Errors: If the translation is perfect.\\
    
    Return exactly the XML template below (no additional tags): If there are multiple error types, provide them as a comma-separated list inside the <error\_type> tag.\\
    <reasoning>\\
      <error\_type> ERROR\_TYPE1, ERROR\_TYPE2 </error\_type>\\
      <description>Provide a detailed explanation of the translation errors here.</description>\\
    </reasoning>\\
    <answer><da\_score>0-100</da\_score></answer>\\ 

\textbf{User:} \\[0.25em] 
\textit{\textbf{Source (English)}}: \{Source Sentence\}\\[0.25em]
\textit{\textbf{Hypothesis (Malayalam)}} : \{Translated Sentence\}\\[0.25em] 

Read the weak human annotations for translated sentences and produce your detailed reasoning, error\_type, description and DA score. \\ [1em] 
\textbf{Assistant:}\\[0.25em] 
\textbf{\{\textit{Weak human annotations}\}}: \{<Remark 1>, <Remark 2>, <Remark 3>\}

\end{Example}
\caption{Prompt template used for all \aloperl experiments when human annotated \tqr is utilized as weak supervision signal }\label{fig:BASE_PROMPT}

\end{figure}

\subsection{ALOPE-RL Framework }\label{subsec:alopeplus}

\aloperl formulates translation quality estimation as a policy optimization problem, using GRPO ~\autocite{shao2024deepseekmathpushinglimitsmathematical} with multi-component rewards to align model predictions with human-assessed translation quality. Unlike direct supervised fine-tuning, GRPO (Group Relative Policy Optimization) leverages \emph{grouped comparisons of multiple candidate generations} to adjust the policy toward outputs that achieve higher task-specific rewards. 
In our approach, the `policy' corresponds to the QE model, indicated by $\pi_\theta$, which generates structured responses containing: (i) A direct assessment score, (ii) Error category(s), and (iii) A natural language explanation of the translation error as outlined in the prompt Figure ~\ref{fig:BASE_PROMPT}).

\begin{figure*}[t]
 \centering
 \includegraphics[width=0.9\textwidth, keepaspectratio]{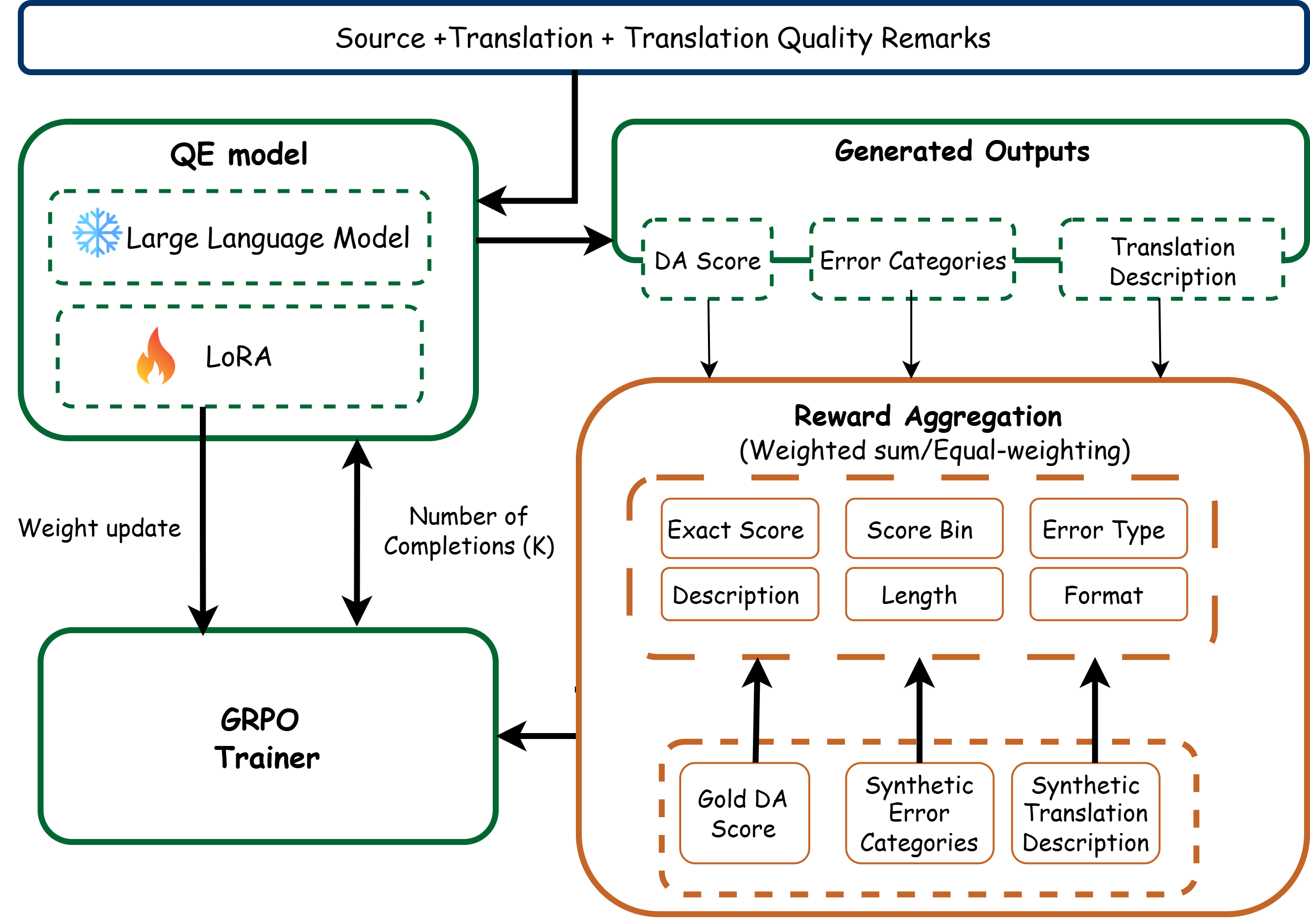}
\caption{Architecture diagram of \aloperl}
\label{fig:alope-rl} 
\end{figure*}

For the given input $x$, we sample a set of $K$ generations $\{y_1, y_2, \dots, y_K\}$ from the policy $\pi_\theta(y \mid x)$. Each generation receives a reward $R(y_i, x)$, computed from multiple components specific to QE:

\begin{itemize}

    \item \textbf{Exact Score Reward ($R_{\text{exact}}$):}  
    To capture coarse alignment, we use a linear penalty:
    \[
    R_{\text{exact}} = 1 - \frac{|s_{\text{pred}} - s_{\text{gold}}|}{100}
    \]\label{eq:exact_score}
    where $s_{\text{pred}}$ is the predicted DA score and $s_{\text{gold}}$ is the ground truth DA score (computed as the average of the three human annotators’ judgments). This gives $1.0$ for exact matches and is reduced linearly with the distance to the ground truth.
    \\
    \item \textbf{Score Bin Reward ($R_{\text{bin}}$):}  
    DA scores are discretized into six bins: $[1,10]$, $[11,30]$, $[31,50]$, $[51,70]$, $[71,90]$, and $[91,100]$. If the predicted and gold scores fall in the same bin, $R_{\text{bin}} = 1.0$. If the prediction differs by $\pm5$ from the gold score, the reward is $0.9$, even if the bins differ. For predictions one bin away, $R_{\text{bin}} = 0.5$. If two bins away and within 15 points of the gold score, the reward is $0.3$; otherwise, it is $0$. This reward acts as a form of reward shaping, encouraging the model to learn the appropriate DA score range for a translation by providing graded feedback for predictions that fall in or near the correct bin.
    \\
    \item \textbf{Error Type Reward ($R_{\text{err}}$):}  
    The predicted and gold annotations each contain a set of error type(s), defined according to the categories specified in the prompt (Figure~\ref{fig:syn_TQR}). To measure their similarity, we compute the Jaccard index~\autocite{article}:        
    \[
    R_{\text{err}} = \frac{|E_{\text{pred}} \cap E_{\text{gold}}|}{|E_{\text{pred}} \cup E_{\text{gold}}|}
    \]
    
    This reward encourages the model to identify relevant translation error types by granting partial credit for overlapping error categories.
    \\
    \item \textbf{Description Reward ($R_{\text{desc}}$):}  
    To assess the semantic quality of the predicted translation description, we use BERTScore F1 ~\autocite{zhang2020bertscoreevaluatingtextgeneration}, which captures the semantic similarity. Here $d_{\text{pred}}$ denotes the predicted description of the translation and the $d_{\text{gold}}$ denotes the synthetic translation description, which we consider as ground truth:
    \[
    R_{\text{desc}} = \text{BERTScore}_{F1}(d_{\text{pred}}, d_{\text{gold}})
    \]

    \item \textbf{Length Reward ($R_{\text{len}}$):}  
    Target for detailed but concise translation description is $\sim$100 words. We reward description lengths in the range 60\text{--}140 words with a full reward (1.0), otherwise applying a soft penalty. The reward is clipped at zero, preventing negative rewards. $L$ represent the length of the predicted description:
        \[
    R_{\text{len}} =
    \begin{cases}
    \begin{array}{@{}l@{}l@{}}
    1 & {{\scriptsize\text{60}}\!\le\!{\scriptsize\text{L}}\!\le\!{\scriptsize\text{140}}} \\
    \max\!\left(0,1-\frac{|L-100|}{100}\right)
    \end{array}
    \end{cases}
    \]
    
    \item \textbf{Format Reward ($R_{\text{fmt}}$):}  
    Outputs must follow a strict XML schema as instructed in the input prompt ( ~\ref{app:experiment_prompts}). If all required tags (\texttt{<reasoning>}, \texttt{<error\_type>}, \texttt{<description>}, \texttt{<answer>}, \texttt{<da\_score>}) are present with non-empty content, $R_{\text{fmt}}=1.0$. If tags are present but incomplete, $R_{\text{fmt}}=0.5$. Otherwise $R_{\text{fmt}}=0$.
\end{itemize}

Each reward is normalized into $[0,1]$, and the reward weighting for each generation is defined as:

\[
    R(y_i, x) = \sum_{j=1}^N w_j \, R_j(y_i, x),
    \]

where $R_j(y_i, x) \in [0,1]$ denotes the $j$\textsuperscript{th} reward component, and $w_j$ is its corresponding weight.
\\
\\
The reward weighting is categorized into two:

\begin{itemize}
    \item \textbf{Equal-weighting:}  \label{equal_weighting}
    All the utilized reward components contribute equally to the final reward. 
    \\
    \item \textbf{Weighted sum:}  
    Different weights $w_j$ are assigned to each component to emphasize certain aspects of the task.

\end{itemize}

GRPO defines the objective in terms of \emph{relative advantages} ($A$) between generations ($K$), which are the candidate outputs within the same group. The relative advantage of generation $y_i$ is given by:

\[
    A(y_i) = R(y_i, x) - \frac{1}{K}\sum_{k=1}^K R(y_k, x)
\]

The policy update is then performed by maximizing the clipped objective:

{\small
\begin{align*}
    L^{\text{GRPO}}(\theta) &= \mathbb{E}_{x, y \sim \pi_\theta} \Bigg[
    \min \Bigg(
    \frac{\pi_\theta(y \mid x)}{\pi_{\theta_{\text{old}}}(y \mid x)} \, A(y),  \qquad \text{clip}\!\left(\frac{\pi_\theta(y \mid x)}{\pi_{\theta_{\text{old}}}(y \mid x)},
       1 - \epsilon, 1 + \epsilon \right) \, A(y)
    \Bigg)
    \Bigg]
\end{align*}
}

$L^{\text{GRPO}}(\theta)$ denotes the GRPO training objective, which is the clipped surrogate loss used to optimize the policy parameters $\theta$. It balances improving the probability of generations with positive relative advantage $A(y)$ against constraining the policy update through ratio clipping, thereby ensuring stable reinforcement learning. In this equation, the ratio $\frac{\pi_\theta}{\pi_{\theta_{\text{old}}}}$ compares the updated and previous policy probabilities, $\epsilon$ is a small constant controlling the clipping range, and $\text{clip}$ restricts the ratio to lie within $[1-\epsilon,\, 1+\epsilon]$. 

The expectation $\mathbb{E}_{x, y \sim \pi_\theta}[\cdot]$ 
denotes averaging over the training samples $x$ and the corresponding 
generations $y$ drawn from the model’s current policy 
$\pi_\theta(y \mid x)$, ensuring that the objective reflects 
the model’s overall behavior across the dataset rather than 
individual examples.
We set the KL penalty which is originally used in GRPO~\autocite{shao2024deepseekmathpushinglimitsmathematical}, to zero to encourage more substantial policy updates and speed up training~\autocite{liu2025understanding}.
\\

We fine-tune the LoRA adapters to ensure parameter-efficiency, updating only low-rank matrices within attention and feedforward layers. During GRPO-based training, the surrogate loss $L^{\text{GRPO}}(\theta)$ is backpropagated only through the LoRA parameters. We sample \textit{4 generations per input} during the training ($K=4$). The trainer uses the rewards to rank generations and align the model parameters. Figure ~\ref{fig:alope-rl} shows the overall architecture of \aloperl.

\begin{table}[t]
\centering
\small
\begin{tabular}{l c c c}
\toprule
\textbf{Synthetic Data} & \textbf{$r$} & \textbf{$\rho$} & \textbf{$\tau$} \\
\midrule
\deepseek & 0.420 & \textbf{0.543} & 0.420 \\
\gemini-2.5-pro   & 0.560 & 0.526 & 0.402 \\
\bottomrule
\end{tabular}
\caption{Comparison of the Spearman correlation scores obtained with \aloperl using synthetic explanations from \deepseek and \gemini-2.5-pro. The highest Spearman correlation is shown in bold.}
\label{tab:synthetic_data_results}
\end{table}

\subsection{Experimental Setup}\label{subsec:experimental-setup}
All the experiments were conducted on single/dual \texttt{NVIDIA A100 GPUs (80GB)}. We evaluated three instruction-tuned LLMs: \qwen, \gemma , \llama \footnote{
\href{https://huggingface.co/google/gemma-3-4b-it}{\gemma-IT}, \href{https://huggingface.co/unsloth/Qwen3-4B-Instruct-2507-unsloth-bnb-4bit/discussions}{\qwen-INSTRUCT}, \href{https://huggingface.co/unsloth/Llama-3.2-3B-Instruct}{\llama-INSTRUCT}}. To ensure computational efficiency, all the selected models had $\leq$4B parameters and were fine-tuned with 4-bit quantization using LoRA adapters. Training was performed for up to 100 optimization steps with a learning rate of $5\times10^{-5}$, batch size of 64 (with gradient accumulation steps set to 4), and LoRA configured with rank $r=32$, $\alpha=64$, and dropout $p=0.05$. LoRA adapters were applied to both attention and feed-forward layers.

\subsection{Evaluation \& Metric}
We primarily evaluate QE performance using Spearman’s rank correlation coefficient ($\rho$)~\autocite{sedgwick2014spearman} between predicted and gold DA scores, as it reflects agreement in relative ranking. Since each segment is annotated by multiple annotators, the mean DA score is used as the gold reference. We additionally report Pearson’s correlation coefficient ($r$)~\autocite{cohen2009pearson} to measure linear correlation and Kendall’s Tau ($\tau$)~\autocite{lapata-2006-automatic} to assess pairwise ranking consistency. Statistical significance between competing methods is assessed using the Williams significance test~\autocite{graham-baldwin-2014-testing,williams1959regression}.

\section{Results \& Discussion}
This section reports the results and findings of \aloperl and baseline experiments on the En$\rightarrow$Ml dataset with \tqr, as well as on additional language pairs with \wtag.
 
\subsection{Optimizing for Explanations} 
Initial experiments on the En$\rightarrow$Ml language pair with \aloperl were conducted to select the most effective source of synthetic explanations for QE, which serve as reference signals for the error category and  translation description based reward calculations. Using the \gemma model with a fixed multi-reward setup and \equalweightedmean aggregation ($\S$~\ref{subsec:alopeplus}), we compare synthetic explanations from \deepseek and \gemini-2.5-pro to isolate the impact of explanation quality. As shown in Table~\ref{tab:synthetic_data_results}, synthetic explanations from \deepseek led to higher QE performance. Consequently, all subsequent experiments use \deepseek-generated synthetic data as the reference signal for reward calculation.

\begin{table*}[t]
\centering
\small
\setlength{\tabcolsep}{4pt}
\renewcommand\arraystretch{1.15}

\begin{tabular}{>{\centering\arraybackslash}p{3cm} *{9}{c}}
\toprule

\makecell[c]{\textbf{Reward}} &
\multicolumn{3}{c}{\textbf{\gemma}} &
\multicolumn{3}{c}{\textbf{\llama}} &
\multicolumn{3}{c}{\textbf{\qwen}} \\

\cmidrule(lr){2-4} \cmidrule(lr){5-7} \cmidrule(lr){8-10}
\multicolumn{1}{c}{\textbf{Config}} &
\(r\) & \(\rho\) & \(\tau\)
& \(r\) & \(\rho\) & \(\tau\)
& \(r\) & \(\rho\) & \(\tau\) \\
\midrule

All rewards
& 0.420 & 0.543 & 0.420
& 0.133 & 0.148 & 0.110
& 0.182 & \textsuperscript{*}0.233 & 0.177 \\
\midrule

Core rewards
& 0.583 & \textsuperscript{*}\textbf{0.571} & 0.436
& 0.265 & 0.283 & 0.213
& 0.157 & \textsuperscript{*}0.225 & 0.170 \\
\midrule

DA only rewards
& 0.585 & 0.550 & 0.428
& 0.238 & \textsuperscript{*}0.292 & 0.214
& 0.234 & \textsuperscript{*}0.263 & 0.197 \\
\bottomrule
\end{tabular}

\caption{Correlation scores (\(r\), \(\rho\), and \(\tau\)) obtained with \aloperl under different reward configurations across models for En~$\rightarrow$Ml. The bold value indicates the highest Spearman correlation score (\(\rho\)) obtained comparing all the models and reward configurations.(\textsuperscript{*}) indicates the winning Spearman based on the weighting strategy of \weightedsum for each model in each reward setting. A detailed table of this can be found in the ~\ref{app:extended_en-ml_table}}
\label{tab:en-ml_qegrpo_results_simplified}
\end{table*}

\subsection{ALOPE-RL with TQR } 
\aloperl experiments were conducted using three models ($\S$~\ref{subsec:experimental-setup}) and examined under three primary reward configurations. Rather than fine-grained re-weighting within a fixed reward set, our study focuses on structural reward ablations, which isolates the contribution of different reward components.

\begin{itemize}
    \item 
\textbf{\allrewards} - Employs all six reward components described in $\S$~\ref{subsec:alopeplus}. The weights are set as w = \texttt{[0.15, 0.25, 0.20, 0.20, 0.10, 0.10]} for $\{R_{\text{bin}},\, R_{\text{exact}},\, R_{\text{err}},\, R_{\text{desc}},\, R_{\text{len}},\, R_{\text{fmt}}\}$. Higher weights are assigned to rewards that directly enforce alignment with DA scores ($R_{\text{bin}}, R_{\text{exact}}$) and for the error-type supervision ($R_{\text{err}}$).

\item
\textbf{\corerewards} - This reward structure is restricted to four key components: $R_{\text{bin}}$, $R_{\text{exact}}$, $R_{\text{err}}$, and $R_{\text{fmt}}$. The corresponding weights are set to \texttt{ w = [0.30, 0.30, 0.25, 0.15]}. This configuration is motivated by the objective to simplify the reward signal and reduce potential noise from $R_{\text{len}}$ and $R_{\text{desc}}$ rewards, which may negatively impact performance.
\\
\item
\textbf{\daonly} - This configuration is solely based on the DA score and format-based rewards. w = \texttt{[0.45, 0.45, 0.10]} for $R_{\text{bin}}$, $R_{\text{exact}}$, $R_{\text{fmt}}$. This reward structure acts as an ablation baseline to evaluate whether DA-based rewards alone are sufficient, or whether additional gains arise from \tqr-derived rewards in other configurations.

\end{itemize}

In addition to the \weightedsum strategy, each reward configuration is also evaluated using the \equalweightedmean (\S~\ref{subsec:alopeplus}). All experiments use prompts augmented with \tqr as weak supervision signals (Figure~\ref{fig:BASE_PROMPT}). Results are reported in Table~\ref{tab:en-ml_qegrpo_results_simplified} and evaluated using Spearman’s correlation on DA score prediction.

Among the evaluated configurations, the \corerewards-only setting combined with the \weightedsum strategy with \gemma achieves the highest Spearman correlation among all three models. In this setting, the error-type reward introduces supervision with error-categories to refine the reasoning for the score and aligns it with human judgments. When all six rewards are employed (\allrewards), the description-based and length-based rewards ($R_{\text{desc}}$, $R_{\text{len}}$) add richer context but also increase the complexity of reward calculation. These components depend on BERTScore similarity for explanations and soft word-length constraints, which introduce variability when models generate free-form explanations. This added noise can diffuse the reward signal, making GRPO updates less aligned with the DA prediction, which explains the drop in correlation compared to the highest correlation of \corerewards. Moreover, the highest performance obtained with \corerewards configuration compared to \daonly indicates that incorporating error-category based reward provides meaningful additional supervision with better reasoning of LLMs for QE beyond DA-only signals.

Overall, \gemma consistently achieves substantially higher QE performance than both \qwen and \llama across all reward configurations. While the latter models tend to perform best under the simpler \daonly setting, suggesting greater sensitivity to noise when auxiliary rewards are introduced \gemma remains robust to richer supervision under GRPO. This highlights \gemma’s stronger capacity to exploit structured reward signals for quality estimation.

\begin{table}[t]
\centering
\small
\setlength{\tabcolsep}{4pt}
\renewcommand{\arraystretch}{1.15}

\begin{tabular}{>{\raggedright\arraybackslash}p{3.0cm}ccc}
\toprule

\makecell[c]{\textbf{Setting}} &
\(r\) & \(\rho\) & \(\tau\) \\
\midrule

Zero-shot\textsuperscript{*}                 
& 0.222 & 0.226\textsuperscript{$\dagger$} & 0.170 \\
Zero-shot with \tqr\textsuperscript{*}       
& 0.435 & 0.411\textsuperscript{$\dagger$} & 0.312 \\
IFT\textsuperscript{*}                       
& 0.295 & 0.277\textsuperscript{$\dagger$} & 0.197 \\
IFT with \tqr\textsuperscript{*}             
& 0.538 & 0.525\textsuperscript{$\dagger$} & 0.382 \\
\deepseek                                    
& 0.475 & 0.474\textsuperscript{$\dagger$} & 0.355 \\
\deepseek with \tqr                           
& 0.566 & 0.562                              & 0.426 \\
\xcomet-XL                                   
& 0.355 & 0.352\textsuperscript{$\dagger$} & 0.245 \\
\cometkiwi                                   
& 0.454 & 0.474\textsuperscript{$\dagger$} & 0.333 \\
\aloperl\textsuperscript{*}                  
& 0.583 & \textbf{0.571}                    & 0.436 \\
\bottomrule
\end{tabular}

\caption{Correlation scores (\(r\), \(\rho\), and \(\tau\)) obtained from the best settings of the baselines, SOTA approaches, and \aloperl for En$\rightarrow$Ml. Settings marked with (*) used \gemma. (\textsuperscript{$\dagger$}) indicates statistically significant Spearman scores compared to the best (bold).}
\label{tab:comparison_en-ml}
\end{table}

We analyzed the impact of different weighting strategies within the \aloperl framework to determine which configuration yields more consistent performance. However, no single weighting method consistently outperformed the other across all models and settings. Notably (Table ~\ref{tab:en-ml_qegrpo_results_simplified}), when \allrewards were utilized, the \equalweightedmean resulted in higher Spearman correlations for the majority of instances, possibly because it normalizes the influence of each reward component, reducing sensitivity to noise introduced by semantically complex signals such as $R_{\text{desc}}$, $R_{\text{len}}$ rewards. In contrast, the \weightedsum strategy amplifies whichever components dominate the reward vector, which yields strong performance only when the high-weight rewards are less noisy and well-aligned with the primary task, evident in the \corerewards and \daonly setting (See ~\ref{app:extended_en-ml_table} for detailed results).

\subsubsection{Comparative Benchmarking} \label{sec:comparitive-benchmarking}
For the comparative analysis, we benchmarked the best-performing configuration of \aloperl against multiple baselines and state-of-the-art (SOTA) QE systems as shown in Table~\ref {tab:comparison_en-ml}. The baselines included the zero-shot and instruction-fine-tuned (IFT) variants of \gemma, as well as zero-shot evaluations using \deepseek\footnote{\href{https://api-docs.deepseek.com/news/news250929}{DeepSeek-V3.2-Exp}}. In addition, we compared our model against two SOTA QE systems-\xcomet \footnote{\href{https://huggingface.co/Unbabel/XCOMET-XL}{Unbabel/XCOMET-XL}} and \cometkiwi\footnote{\href{https://huggingface.co/Unbabel/wmt23-cometkiwi-da-xl}{Unbabel/wmt23-cometkiwi-da-xl}}, which represent the current leading encoder-based approaches. All the LLM-based results (Table ~\ref{tab:comparison_en-ml}) are obtained utilizing the similar prompt (Figure~\ref{fig:BASE_PROMPT}), with and without \tqr, and employ \gemma as the base model, given its superior performance in prior \aloperl experiments (Table~\ref{tab:en-ml_qegrpo_results_simplified}).  The IFT results are obtained by instruction fine-tuning the model using the mean DA scores together with synthetic explanations (comprising error categories and translation descriptions), which are treated as reference signals during training. We additionally evaluate \deepseek as a high-capacity LLM baseline due to its strong multilingual reasoning capabilities~\autocite{deepseekai2025deepseekv3technicalreport}.

\cometkiwi ~\autocite{rei-etal-2023-scaling} was built using a large-scale training corpus containing more than $940,000$ examples across $38$ language pairs. This corpus brings together a significant amount of data: Direct Assessment annotations from WMT editions (2017–2020), the MLQE-PE dataset~\autocite{fomicheva-etal-2022-mlqe}, as well as resources from the 2023 QE shared task~\autocite{blain-etal-2023-findings}. Following this broad pretraining, the model undergoes an additional fine-tuning stage specifically targeting sentence-level quality estimation~\autocite{rei-etal-2023-scaling}. \xcomet~\autocite{guerreiro-etal-2024-xcomet} extends on this using large-scale MQM-annotated data.

As shown in the Table~\ref {tab:comparison_en-ml}, our proposed \aloperl approach has obtained the new SOTA performance, outperforming all baselines and prior SOTA systems. The improvement is statistically significant over all compared methods, except for \deepseek with \tqr. Notably, \aloperl attains this performance using only 4K training instances and compact models ($\leq$4B parameters) in a quantized setting, highlighting the efficiency of the proposed approach. This observation is further supported by our ablation study (~\ref{app:ablation_data_count}), where \aloperl maintains higher correlation scores even with reduced training data.

\begin{table*}[t]
\centering
\small
\setlength{\tabcolsep}{4pt}
\renewcommand\arraystretch{1.15}

\begin{tabular}{>{\raggedright\arraybackslash}p{1.47cm} *{12}{c}}
\toprule

\makecell[c]{\textbf{Setting}} &
\multicolumn{3}{c}{\textbf{En$\rightarrow$Mr}} &
\multicolumn{3}{c}{\textbf{En$\rightarrow$Hi}} &
\multicolumn{3}{c}{\textbf{En$\rightarrow$Ta}} &
\multicolumn{3}{c}{\textbf{En$\rightarrow$Ml}} \\

\cmidrule(lr){2-4} \cmidrule(lr){5-7} \cmidrule(lr){8-10} \cmidrule(lr){11-13}
\multicolumn{1}{c}{} &
\(r\) & \(\rho\) & \(\tau\)
& \(r\) & \(\rho\) & \(\tau\)
& \(r\) & \(\rho\) & \(\tau\)
& \(r\) & \(\rho\) & \(\tau\) \\
\midrule

\makecell[l]{\textbf{\aloperl}\\w/ \wtag}
& 0.246 & \textsuperscript{*}0.249 & 0.198
& 0.233 & 0.122 & 0.093
& 0.294 & \textsuperscript{*}\textbf{0.311} & 0.242
& 0.334 & \textsuperscript{*}\textbf{0.331} & 0.254 \\
\midrule

\makecell[l]{\textbf{\aloperl}\\w/o \wtag}
& 0.273 & \textbf{0.269} & 0.200
& 0.289 & \textsuperscript{*}\textbf{0.179} & 0.142
& 0.318 & \textsuperscript{*}0.277 & 0.207
& 0.233 & \textsuperscript{*}0.241 & 0.176 \\
\midrule

IFT
& 0.265 & 0.247 & 0.173
& 0.159 & 0.114 & 0.080
& 0.297 & 0.288 & 0.200
& 0.295 & 0.277 & 0.197 \\
\bottomrule

\end{tabular}

\caption{Correlation scores (\(r\), \(\rho\), and \(\tau\)) obtained from \aloperl and instruction fine-tuned (IFT) models across multiple language pairs in different settings. Bold values indicate the highest Spearman correlation for each language pair. Spearman scores marked with (\textsuperscript{*}) indicate the better scores obtained with \corerewards configuration for each language pair and setting. An extended table with all the scores is available in App.~\ref{app:Otherlang_table_extended}.}
\label{tab:other_lang_pair_results}
\end{table*}

\subsection{Other language-pairs}  
For the experiments on En~$\rightarrow$Mr, En~$\rightarrow$Hi, and En~$\rightarrow$Ta, we selected only the \texttt{\gemma} model along with the best-performing reward weighting strategies identified for the En~$\rightarrow$Ml setup: the \weightedsum strategy for the \corerewards configuration and the \equalweightedmean for the \allrewards configuration. The \aloperl experiments employed \wtag ($\S$~\ref{other_lang_synthetic_data}) as weak supervision signals, replacing the \tqr annotations used in earlier experiments (~\ref{app:experiment_prompts}- Figure~\ref{app:rlqe_word_level_prompt}).  To further assess the contribution of these weak supervision signals, we also conducted ablation experiments that excluded them. All results were compared against a baseline obtained through IFT of the same \texttt{\gemma} model and prompt, as reported in Table~\ref{app:Otherlang_table_extended}. We also evaluate En$\rightarrow$Ml under the \wtag setting.

The results demonstrate that \aloperl consistently outperforms the IFT baseline across all evaluated language pairs, underscoring the robustness and adaptability of our framework. Incorporating \wtag as weak supervision signals yielded performance gains for only two language pairs—En$\rightarrow$Ml and En$\rightarrow$Ta, both belonging to the Dravidian language family. In contrast, for the Indo-Aryan pairs (En$\rightarrow$Mr and En$\rightarrow$Hi), models trained without \wtag achieved comparatively better performance. This discrepancy may be attributed to morphological and syntactic differences between the language families. Dravidian languages exhibit richer inflectional morphology and agglutinative structures, where token-level cues can more effectively capture localized translation errors (\textit{i.e} suffixal inflections and compounding). In contrast, Indo-Aryan languages tend to have more on word order and context-dependent grammatical constructions, where token-level alignment signals may fail to reflect semantic or structural errors at the word-level. Consequently, \wtag introduces noise rather than guidance for RL for such language pairs.

\begin{figure}[t]
    \centering
    \includegraphics[width=0.85\columnwidth]{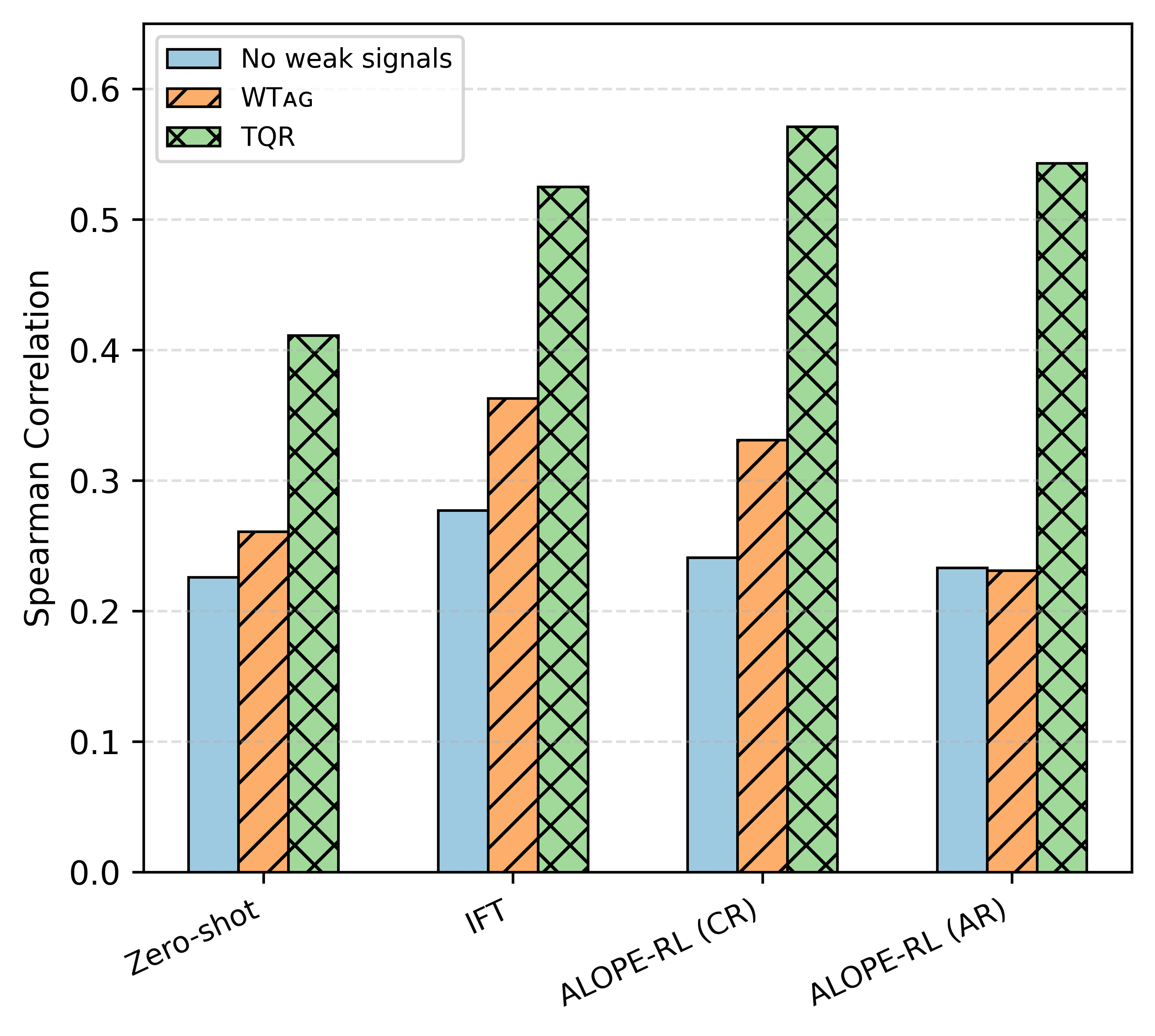}
    \caption{Performance comparison on En$\rightarrow$Ml across different weak-signal settings. IFT = Instruction Fine-Tuning; CR = \corerewards; AR = \allrewards.}
    \label{fig:compare_en_ml}
\end{figure}

\subsection{\tqr as Contextual Weak Signal} 

Figure~\ref{fig:compare_en_ml} presents a comparative analysis on the En$\rightarrow$Ml language pair to examine how LLM performance varies with different weak supervision signal configurations: \tqr, \wtag, and the absence of any weak signals. We evaluate these signals across zero-shot, IFT, and \aloperl (with both \corerewards and \allrewards). 
The correlation results show that incorporating \tqr consistently yields substantially higher Spearman correlations than using \wtag or with no weak signals in any of the experimental settings. This confirms the effectiveness of \tqr as a complementary weak supervision signal to DA score prediction with LLMs. Furthermore, the results demonstrate that the inclusion of \tqr achieves the best performance with the \aloperl framework, underscoring the effectiveness of our proposed approach.

Additionally, we investigated the Mean Squared Error (MSE) and Mean Absolute Error (MAE) values obtained for En$\rightarrow$Ml with \aloperl as shown in Table~\ref{tab:alope_rl_wordtag_tqr}. When \tqr signals were available, absolute RL rewards (\textit{e.g.,} exact DA and bin scores) consistently yielded strong gains, leading to SOTA correlation. In contrast, models trained only with \wtag showed weaker correlation, despite reasonable absolute losses (MSE/MAE), likely because \tqr implicitly encodes a preference structure that our error-based rewards can utilize, whereas \wtag lacks such details, and the RL objective remains purely absolute. Addressing this limitation may require incorporating ranking-based supervision (\textit{e.g.,} correlation-driven rewards), but such approaches demand substantially more data and are therefore left for future work.

In all experiments, the model predicts error categories and translation descriptions in addition to DA scores. We report the performance of these additional predictions in~\ref{app:error_cat}; however, since our primary objective is improving segment-level QE, a detailed analysis of these outputs is left for future work.

\begin{table}[t]
\centering
\small
\setlength{\tabcolsep}{4pt}
\renewcommand{\arraystretch}{1.15}

\begin{tabular}{>{\raggedright\arraybackslash}p{2.5cm}cccc}
\toprule
\textbf{ALOPE-RL Setting} & \textbf{Reward Config.} & \textbf{MSE} & \textbf{MAE} & \boldmath{$\rho$} \\
\midrule

\multirow[t]{2}{*}{\cellcolor{tbrowcolor}\parbox[t]{3.0cm}{\rule{0pt}{2.6ex}With \tqr}}
& \allrewards  & 393.412 & 14.70 & 0.543 \\
& \corerewards & 371.913 & 14.73 & 0.571 \\
\midrule

\multirow[t]{2}{*}{\cellcolor{tbrowcolor}\parbox[t]{3.0cm}{\rule{0pt}{2.6ex}With \wtag}}
& \allrewards  & 577.677 & 19.77 & 0.231 \\
& \corerewards & 494.363 & 18.73 & 0.331 \\
\bottomrule
\end{tabular}

\caption{MAE, MSE and Spearman correlation values obtained with \aloperl under different type of weak signals (\tqr and \wtag) and reward configurations (\allrewards, \corerewards).}
\label{tab:alope_rl_wordtag_tqr}
\end{table}

\section{Conclusion}
This work advances reference-less machine translation evaluation by demonstrating that lightweight, error-aware supervision can substantially improve LLM-based QE in low-resource settings. By introducing the first segment-level English→Malayalam QE dataset enriched with Translation Quality Remarks (TQR), we show that short, free-form annotator comments provide a practical middle ground between scalar DA scores and costly fine-grained error annotations. Unlike traditional QE approach that collapse quality into a single numeric signal, TQR preserves contextual information about translation errors while remaining scalable and annotation-efficient.

Building on this dataset, we proposed ALOPE-RL, a policy-based reinforcement learning framework that aligns LLM predictions with human judgment through multi-component rewards capturing both score accuracy and error awareness. Our experiments demonstrate that this approach enables compact, quantized LLMs to outperform strong encoder-based QE systems and larger LLM baselines, despite being trained on substantially fewer annotated instances. The results further indicate that error-category–driven rewards play a crucial role in improving correlation with human judgments, highlighting the importance of modeling why a translation is incorrect rather than only how much it deviates from the source.

Our experiments further show that the choice of weak supervision signal is critical. In particular, tokenized word-level error tags provide limited benefits compared to Translation Quality Remarks, which consistently yield stronger gains. These findings indicate that natural-language error remarks constitute a more flexible and language-agnostic form of weak supervision for quality estimation than rigid token-level annotations.

Overall, this study provides empirical evidence that error-aware, policy-optimized learning is a viable and efficient pathway for QE under limited data and compute budgets. Future work will extend this framework by conducting a deeper analysis of the predicted error categories and translation descriptions, and exploring correlation-aware reward formulations.

\paragraph{Acknowledgments}  We are grateful for the native Malayalam language speakers, Silpa Vadakkeeveetil Sreelatha and Adarsh Kappiyath, who have helped us to validate the prompt outputs. We also want to thank Sweta Agrawal for her guidance on this study. Also, we would like to thank the European Association of Machine Translation (EAMT) for supporting the creation of the English$\rightarrow$Malayalam dataset curation. 






\defbibnote{preamble}

\printbibliography[prenote={preamble}]

\clearpage
\appendix

\section{Annotation of English to Malayalam  QE Dataset} \label{app:DA_guideline}
The annotators were provided with detailed guidelines outlining the Direct Assessment (DA) scoring criteria and examples of common error types. To build the dataset, we curated parallel segments from publicly available corpora of Anuvaad spanning across diverse domains (News, Finance, Legal). Given that the Anuvaad corpus contains substantial noise, we applied LaBSE ~\autocite{feng-etal-2022-language} similarity filtering, retaining only parallel sentence pairs above a high-quality threshold ($\geq 0.80$)
. This helped guarantee the availability of reasonably accurate reference translations for downstream comparison. To obtain a balanced distribution of sentence lengths while controlling annotation costs, we sampled source sentences in predefined token-length ranges (0–10, 10–20, 20–30 tokens). Target translations were generated using the IndicTrans2~\autocite{gala2023indictrans2highqualityaccessiblemachine} MT model. Throughout the annotation process, a native speaker not involved in the main annotation effort periodically reviewed a random subset of samples each week. Their feedback was incorporated to improve annotation consistency and quality. After all three annotators completed the DA scoring, we computed inter-annotator agreement by calculating pairwise Pearson and Spearman correlations for all annotator pairs (A1–A2, A1–A3, and A2–A3) based on their DA scores. The final agreement for each metric is obtained by averaging the three pairwise correlations, yielding an average Pearson correlation of 0.952 and an average Spearman correlation of 0.924, indicating very strong consistency in both absolute scoring and relative ranking across annotators. Finally, we split the data into training and test splits, ensuring that the distribution of DA scores remained balanced across the sets.

\subsection{Translation Quality Remarks (TQR)} \label{app:tqr-guideline}

To complement the DA scores, annotators were guided to provide concise Translation Quality Remarks (TQR) describing any errors observed in the machine-translated output. Annotators were asked to keep remarks brief, explicitly identify the problematic word(s), and specify the corresponding error type(s) from a predefined taxonomy: Untranslated, Addition, Mistranslation, Fluency Error, or Other. These error categories broadly align with the \href{https://themqm.org/error-types-2/typology/}{Multi-Dimensional Quality Metric Error Taxonomy} (MQM;~\autocite{lommel-etal-2014-using}) and are based on findings for low-resource quality estimation~\autocite{sindhujan2024optimizing}. For each identified error, annotators were encouraged to indicate the erroneous word/words and the corresponding error category, using the given example formats as guidance but annotators \textit{were given the freedom} to adopt any phrasing or structure they found natural, as long as the error was clearly indicated. Also, annotators had the flexibility to omit remarks when a translation was considered accurate and contained no identifiable issues, resulting in some instances without an associated TQR.

\clearpage

\section{Results obtained with ALOPE-RL for En~$\rightarrow$Ml with TQR} \label{app:extended_en-ml_table}

\begin{table*}[ht]
\small
\setlength{\tabcolsep}{3pt}
\renewcommand\arraystretch{1.15}

\begin{tabularx}{\textwidth}{
>{\centering\arraybackslash}p{2.1cm}
>{\centering\arraybackslash}p{2.2cm}
*{9}{>{\centering\arraybackslash}X}
}
\toprule

\multirow[t]{2}{*}{\parbox[t]{2.1cm}{\centering\rule{0pt}{3.0ex}\textbf{Reward Config}}} &
\multirow[t]{2}{*}{\parbox[t]{2.2cm}{\centering\rule{0pt}{3.0ex}\textbf{Weighting}\\\textbf{}}} &
\multicolumn{3}{c}{\textbf{\gemma}} &
\multicolumn{3}{c}{\textbf{\llama}} &
\multicolumn{3}{c}{\textbf{\qwen}} \\

\cmidrule(lr){3-5} \cmidrule(lr){6-8} \cmidrule(lr){9-11}
& & \(r\) & \(\rho\) & \(\tau\)
& \(r\) & \(\rho\) & \(\tau\)
& \(r\) & \(\rho\) & \(\tau\) \\
\midrule

\multirow[t]{2}{*}{\cellcolor{tbrowcolor}\parbox[t]{2.1cm}{\centering\rule{0pt}{2.8ex}\textbf{All rewards}}}
& Weighted sum
& 0.502 & 0.505 & 0.382
& 0.050 & $-$0.020 & $-$0.017
& 0.182 & 0.233 & 0.177 \\
& Equal-weighting
& 0.420 & 0.543 & 0.420
& 0.133 & 0.148 & 0.110
& 0.122 & 0.162 & 0.119 \\
\midrule

\multirow[t]{2}{*}{\cellcolor{tbrowcolor}\parbox[t]{2.1cm}{\centering\rule{0pt}{2.8ex}\textbf{Core rewards only}}}
& Weighted sum
& 0.583 & 0.571 & 0.436
& 0.079 & 0.185 & 0.135
& 0.157 & 0.225 & 0.170 \\
& Equal-weighting
& 0.538 & 0.542 & 0.415
& 0.265 & 0.283 & 0.213
& 0.136 & 0.106 & 0.080 \\
\midrule

\multirow[t]{2}{*}{\cellcolor{tbrowcolor}\parbox[t]{2.1cm}{\centering\rule{0pt}{2.8ex}\textbf{DA-only rewards}}}
& Weighted sum
& 0.512 & 0.485 & 0.380
& 0.238 & 0.292 & 0.214
& 0.234 & 0.263 & 0.197 \\
& Equal-weighting
& 0.585 & 0.550 & 0.428
& 0.202 & 0.208 & 0.154
& 0.207 & 0.202 & 0.150 \\
\bottomrule
\end{tabularx}

\caption{Correlation scores (\(r\), \(\rho\), \(\tau\)) obtained with \aloperl for different rewards and weighting strategies across models for En~$\rightarrow$Ml.}
\label{tab:en-ml_qegrpo_results}
\end{table*}

\section{Results obtained with word-level tags as weak signals}\label{app:Otherlang_table_extended}

\begin{table*}[ht]
\centering
\small
\setlength{\tabcolsep}{3pt}
\renewcommand\arraystretch{1.15}

\begin{tabularx}{\textwidth}{
>{\raggedright\arraybackslash}p{3.3cm}
*{12}{>{\centering\arraybackslash}X}
}
\toprule

\multirow[t]{2}{*}{\parbox[t]{3.3cm}{\raggedright\rule{0pt}{3.0ex}\textbf{Setting}}} &
\multicolumn{3}{c}{\textbf{En~$\rightarrow$Mr}} &
\multicolumn{3}{c}{\textbf{En~$\rightarrow$Hi}} &
\multicolumn{3}{c}{\textbf{En~$\rightarrow$Ta}} &
\multicolumn{3}{c}{\textbf{En~$\rightarrow$Ml}} \\

\cmidrule(lr){2-4}\cmidrule(lr){5-7}\cmidrule(lr){8-10}\cmidrule(lr){11-13}
& \(r\) & \(\rho\) & \(\tau\)
& \(r\) & \(\rho\) & \(\tau\)
& \(r\) & \(\rho\) & \(\tau\)
& \(r\) & \(\rho\) & \(\tau\) \\
\midrule

\cellcolor{tbrowcolor}\parbox[t]{3.3cm}{\raggedright\rule{0pt}{2.8ex}\textbf{\aloperl~(CR)}\\w/ \wtag}
& 0.246 & 0.249 & 0.198
& 0.256 & 0.065 & 0.049
& 0.294 & \textbf{0.311} & 0.242
& 0.334 & \textbf{0.331} & 0.254 \\
\midrule

\cellcolor{tbrowcolor}\parbox[t]{3.3cm}{\raggedright\rule{0pt}{2.8ex}\textbf{\aloperl~(AR)}\\w/ \wtag}
& 0.208 & 0.192 & 0.158
& 0.233 & 0.122 & 0.093
& 0.281 & 0.256 & 0.198
& 0.231 & 0.231 & 0.175 \\
\midrule

\cellcolor{tbrowcolor}\parbox[t]{3.3cm}{\raggedright\rule{0pt}{2.8ex}\textbf{\aloperl~(CR)}\\w/o weak signal}
& 0.262 & 0.265 & 0.199
& 0.289 & \textbf{0.179} & 0.142
& 0.318 & 0.277 & 0.207
& 0.233 & 0.241 & 0.176 \\
\midrule

\cellcolor{tbrowcolor}\parbox[t]{3.3cm}{\raggedright\rule{0pt}{2.8ex}\textbf{\aloperl~(AR)}\\w/o weak signal}
& 0.273 & \textbf{0.269} & 0.200
& 0.277 & 0.145 & 0.113
& 0.285 & 0.213 & 0.155
& 0.222 & 0.233 & 0.172 \\
\midrule

\cellcolor{tbrowcolor}\parbox[t]{3.3cm}{\raggedright\rule{0pt}{2.8ex}\textbf{IFT}}
& 0.265 & 0.247 & 0.173
& 0.159 & 0.114 & 0.080
& 0.297 & 0.288 & 0.200
& 0.295 & 0.277 & 0.197 \\

\bottomrule
\end{tabularx}

\caption{Correlation scores (\(r\), \(\rho\), and \(\tau\)) for \aloperl and IFT across multiple language pairs. Bold values indicate the highest Spearman correlation for each language pair.}
\label{tab:other_lang_pair_results}
\end{table*}
\clearpage


\section{Performance scaling of ALOPE-RL with training instances} \label{app:ablation_data_count}
\begin{figure}[ht]
 \centering
 \includegraphics[width=0.8\textwidth, keepaspectratio]{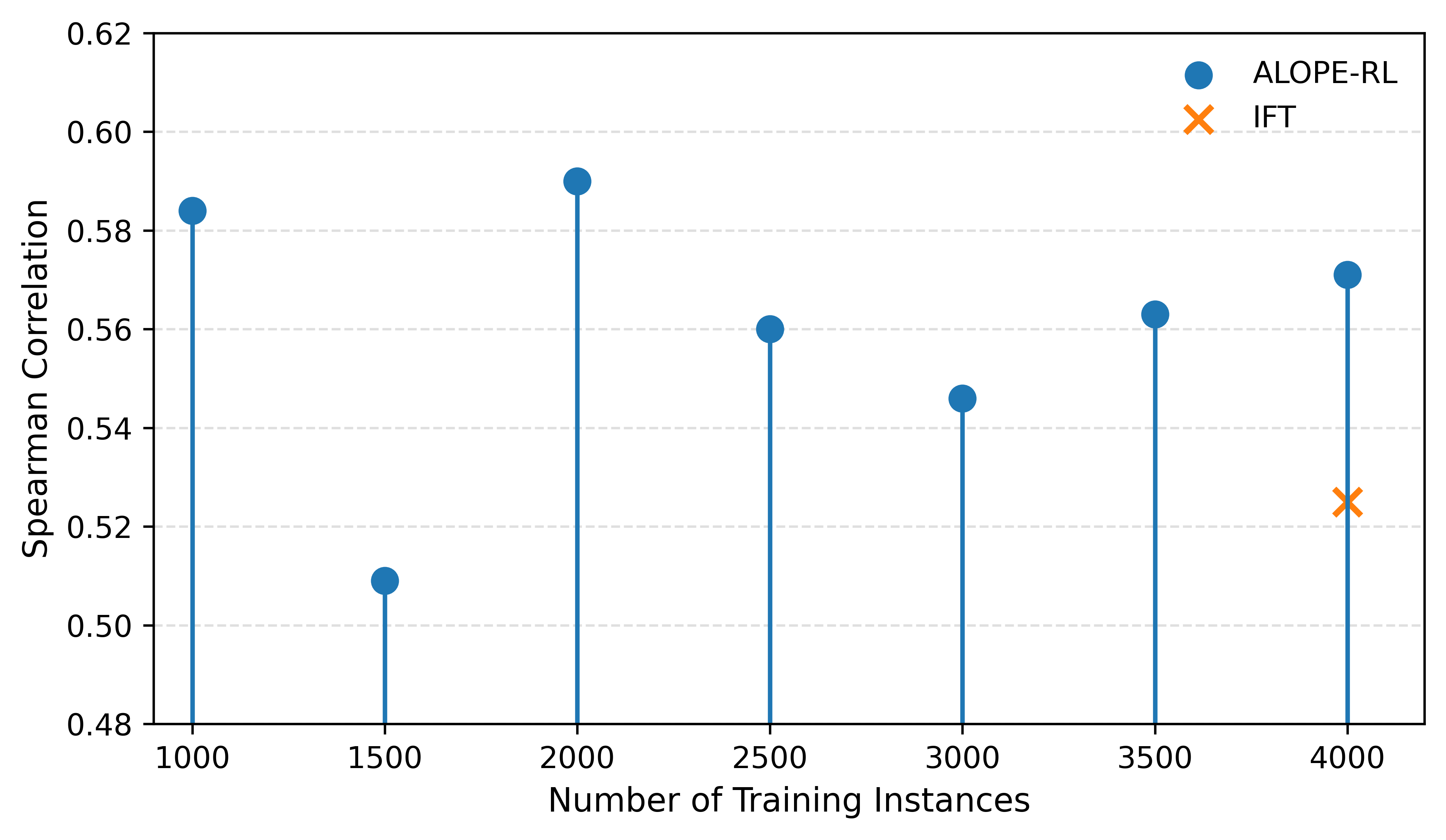}
\caption{ Spearman correlation scores obtained with different number of training instances using \aloperl for En~$\rightarrow$Ml. The plot shows that \aloperl consistently outperforms the IFT (Instruction Fine-Tuning) baseline, even when trained on fewer examples.}
\label{fig:ablation_data_count} 
\end{figure}
\vspace{-10pt}

\section{Results of additional predictions of Error Catgories and Translation Description} \label{app:error_cat}

\begin{table}[H]
\centering
\small
\setlength{\tabcolsep}{3pt}
\renewcommand{\arraystretch}{1.15}

\begin{tabularx}{\textwidth}{
>{\raggedright\arraybackslash}p{2.2cm}
*{8}{>{\centering\arraybackslash}X}
}
\toprule

& \multicolumn{4}{c}{\textbf{Core-rewards only}} 
& \multicolumn{4}{c}{\textbf{All-rewards}} \\
\cmidrule(lr){2-5} \cmidrule(lr){6-9}

\textbf{Error category}
& \textbf{Precision} & \textbf{Recall} & \textbf{F1} & \textbf{Support}
& \textbf{Precision} & \textbf{Recall} & \textbf{F1} & \textbf{Support} \\
\midrule

Untranslated
& 0.616 & 0.730 & 0.668 & 392
& 0.543 & 0.753 & 0.631 & 392 \\

Addition
& 0.949 & 0.528 & 0.679 & 388
& 0.861 & 0.557 & 0.676 & 388 \\

Mistranslation
& 0.739 & 0.719 & 0.729 & 634
& 0.770 & 0.659 & 0.710 & 634 \\

Fluency error
& 0.568 & 0.591 & 0.579 & 411
& 0.586 & 0.455 & 0.512 & 411 \\

Other
& 0.000 & 0.000 & 0.000 & 0
& 0.000 & 0.000 & 0.000 & 0 \\

No errors
& 0.247 & 0.583 & 0.347 & 72
& 0.273 & 0.486 & 0.350 & 72 \\

\midrule
\textbf{Micro avg.}
& 0.637 & 0.649 & 0.643 & 1897
& 0.636 & 0.607 & 0.621 & 1897 \\

\textbf{Macro avg.}
& 0.520 & 0.525 & 0.500 & 1897
& 0.506 & 0.485 & 0.480 & 1897 \\

\midrule
\textbf{BERTScore}
& -- & -- & -- & --
& 0.875 & 0.809 & 0.839 & -- \\
\bottomrule
\end{tabularx}

\caption{The results show the values obtained for the classification of the translation error categories and BERTScore metrics for translation descriptions under different reward configurations.}
\label{tab:error_category_comparison}
\end{table}

\clearpage

\section{Prompt for synthetic data generation with \wtag} \label{app:synthetic_prompts}

\begin{figure}[H]
\centering
\scriptsize
\begin{Example}{} 
You are an expert in identifying the errors in the translations of English to <TARGET LANGUAGE>. The target language is morphologically complex, and in most cases the errors in the machine-translated target segment can be categorized into the following categories:\\

    \textit{Untranslated}: A word or phrase in the source is omitted in the translation.\\
    \textit{Addition}: The translation includes a word or phrase not present in the source.\\
    \textit{Mistranslation}: A word or phrase in the translation does not accurately represent the source meaning.\\
    \textit{Fluency Error}: The translation sounds unnatural due to grammar, spelling, punctuation, or inconsistency.\\
    \textit{Other}: Any other error not covered by the above categories.\\
If there are no errors in the translation, assign the error category as ‘\textit{No Errors}’.\\
    
Your task is to provide a detailed natural language description about the translation discussing any errors within the target segment in 100 words. Use the error categories above and indicate all errors in the translation. \\
The English source segment and the <TARGET LANGUAGE> target segment are provided, along with translation quality remarks which is used to detect word-level errors in the translation. The translated sentence is annotated with token level-tags which has ‘OK’ as a label for tokens translated correctly and ‘BAD’ otherwise.
These token level-tags should be utilized to identify the errors in the translation. For the `Untranslated’ error, the token on the right side of the untranslated text in the translation is annotated with `BAD' tag. 
An additional <EOS> token is appended at the end of the translation segment to account for ‘Untranslated’ error at the end of the segment. If the translation contains no errors and all token-level tags are labeled as `OK', classify the translation under the error category ‘No Errors’. \\

The English source segment and the <TARGET LANGUAGE> target segment need to be analysed are provided below along with the remarks.\\

\textbf{Source segment}: <Source sentence> \\
\textbf{Target segment}: <Tokenized translated sentence with `EOS' token >\\
\textbf{Translation Quality Remark}- \\
\textbf{Token-level tags:} <Tag1, Tag2, ...,Tag N>
\\
Analyse the translation according to the given instructions and return the description of the translation in the following JSON format only:\\
\{\{\\
  "\textbf{Description of the translation}": "",\\
  "\textbf{Identified error categories}": "",\\
\}\}

\end{Example}

\caption{Prompt utilized for the synthetic data generation of language pairs using tokenized word-level tags (\wtag) instead of human-annotated natural language remarks.}\label{fig:syn_wor-level}

\end{figure}

\clearpage
\section{Prompt for the \wtag based experiments}\label{app:experiment_prompts}
\vspace{-10pt}

\begin{figure}[H]
\centering
\scriptsize
\begin{Example}{} \label{otherlang_prompt}

\textbf{System:} \\
You are an expert in evaluating English to <TARGET LANGUAGE> machine translations. Your task is to provide a comprehensive evaluation, including a quality score, error categorization, and a detailed analysis.\\
    \textbf{\textit{Scoring Guidelines (0-100):}}\\
    1.  0-30: Mostly unintelligible - completely inaccurate or containing only some keywords.\\
    2.  31-50: Partial intelligibility - some keywords present but numerous grammatical errors. \\
    3. 51-70: Generally clear - most keywords included with only minor grammatical errors. \\
    4. 71-90: Clear and intelligible - all keywords present with only minor non-grammatical issues. \\
    5. 91-100: Perfect or near-perfect - accurately conveys source meaning without errors. \\
    \\
    \textbf{\textit{Error Categorization Guidelines:}}\\
    1.  Untranslated: A word or phrase in the source is omitted in the translation.\\
    2.  Addition: The translation includes a word or phrase not present in the source.\\
    3.  Mistranslation: A word or phrase in the translation does not accurately represent the source meaning.\\
    4.  Fluency Error: The translation sounds unnatural due to grammar, spelling, punctuation, or inconsistency.\\
    5.  Other: Any other error not covered by the above categories.\\
    6.  No Errors: If the translation is perfect.\\
    
    Return exactly the XML template below (no additional tags): If there are multiple error types, provide them as a comma-separated list inside the <error\_type> tag.\\
    <reasoning>\\
      <error\_type> ERROR\_TYPE1, ERROR\_TYPE2 </error\_type>\\
      <description>Provide a detailed explanation of the translation errors here.</description>\\
    </reasoning>\\
    <answer><da\_score>0-100</da\_score></answer>\\

\textbf{User:} \\[0.25em] 
\textit{\textbf{Source (English)}}: \{Source Sentence\}\\[0.25em]
\textit{\textbf{Hypothesis (<TARGET LANGUAGE>)}} : \{Tokenized translated sentence with `EOS' token\}\\[0.25em] 

There are token-level tags annotated for the translated sentence, which has ‘OK’ as a label for tokens translated correctly and ‘BAD’ otherwise. These token level-tags should be utilized to identify the errors in the translation. For the ‘Untranslated’ error, the token on the right side of the untranslated text in the translation is annotated with 'BAD' tag. An additional <EOS> token is appended at the end of the translation segment to account for ‘Untranslated’ error at the end of the segment.\\ 

Read the Token-level annotations for translated sentences and produce your detailed reasoning, error\_type, description and DA score. \\ [1em] 
\textbf{Assistant:}\\[0.25em] 
\textbf{\{\textit{Token-level annotations}\}}: \{<Tag1, Tag2, ...,Tag N> \}

\end{Example}
\caption{Prompt template used for the \aloperl experiments when tokenized word-level QE tags are utilized as the weak supervision signal.}\label{app:rlqe_word_level_prompt}
\end{figure}

\end{document}